\documentclass[sigconf, nonacm]{acmart}

\newcommand\vldbdoi{XX.XX/XXX.XX}
\newcommand\vldbpages{XXX-XXX}
\newcommand\vldbvolume{18}
\newcommand\vldbissue{11}
\newcommand\vldbyear{2025}
\newcommand\vldbauthors{\authors}
\newcommand\vldbtitle{\shorttitle} 
\newcommand\vldbavailabilityurl{https://github.com/benzhaotang/XXXXX}
\newcommand\vldbpagestyle{empty}

\usepackage{multirow} 
\usepackage{subfigure}
\usepackage{subcaption}
\usepackage{tikz}
\usepackage{pgfplots}
\usepackage{color}
\usepackage{xcolor}
\usepackage{graphicx}
\usetikzlibrary{patterns}
\usepackage{siunitx}
\usepackage[normalem]{ulem}
\useunder{\uline}{\ul}{}
\usepackage{hyperref}
\usepackage{bm} 
\usepackage{colortbl}
\usepackage{enumitem}

\definecolor{c1}{RGB}{42,99,172} %
\definecolor{c2}{RGB}{255,88,93}
\definecolor{c3}{RGB}{255,181,73}
\definecolor{c4}{RGB}{119,71,64} %
\definecolor{c5}{RGB}{228,123,121} %
\definecolor{c6}{RGB}{208,167,39} %
\definecolor{c7}{RGB}{0,51,153}
\definecolor{c8}{RGB}{56,140,139} 
\definecolor{c9}{RGB}{0,0,0} 
\definecolor{color1}{HTML}{FFF4D8}
\definecolor{color2}{HTML}{FF7163}

\definecolor{c10}{RGB}{102,205,170} 
\definecolor{c11}{RGB}{255,149,0}    
\definecolor{c12}{RGB}{173,216,230}  
\definecolor{c13}{RGB}{255,204,0}    
\definecolor{c14}{RGB}{153,0,153}    
\definecolor{c15}{RGB}{255,105,180}  

\definecolor{c16}{RGB}{75,192,192}    
\definecolor{c17}{RGB}{255,215,0}     
\definecolor{c18}{RGB}{54,162,235}    
\definecolor{c19}{RGB}{255,192,203}   
\definecolor{c20}{RGB}{144,238,144}   
\definecolor{c21}{RGB}{240,128,128}   
\definecolor{c22}{RGB}{30,144,255}    
\definecolor{c23}{RGB}{255,255,224}   
\definecolor{c24}{RGB}{220,20,60}     
\definecolor{c25}{RGB}{255,192,203}   
\definecolor{c26}{RGB}{0,158,115}     
\definecolor{Blood}{HTML}{860309}
\definecolor{Olive}{HTML}{807805}
\definecolor{Topaz}{HTML}{F5C678}
\definecolor{American Yellow}{HTML}{F3AA07}
\definecolor{Giants Orange}{HTML}{FF5C1E}
\definecolor{Persian Plum}{HTML}{6C1D2A}
\definecolor{Pearl Aqua}{HTML}{85D6B2}
\definecolor{Moonstone}{HTML}{3F9EBD}
\definecolor{Blue Jeans}{HTML}{5BBCF0}
\definecolor{St. Patrick's Blue}{HTML}{2B2D7C}
\definecolor{Green}{HTML}{159879}
\definecolor{Green2}{HTML}{31D683}
\definecolor{Purple}{HTML}{C1689C}

\newcommand{\MYLOGNAME}{CLAD}

\pgfdeclarepatternformonly{myDenseLines}{\pgfqpoint{-1pt}{-1pt}}{\pgfqpoint{2pt}{2pt}}{\pgfqpoint{1.8pt}{1.8pt}}%
{
  \pgfsetlinewidth{0.8pt} 
  \pgfpathmoveto{\pgfqpoint{0pt}{0pt}}
  \pgfpathlineto{\pgfqpoint{2pt}{2pt}}
  \pgfusepath{stroke}
}

\pgfdeclarepatternformonly{myDenseLinesRev}
  {\pgfqpoint{-1pt}{-1pt}}
  {\pgfqpoint{2pt}{2pt}}
  {\pgfqpoint{1.8pt}{1.8pt}}
  {
    \pgfsetlinewidth{0.8pt}
    \pgfpathmoveto{\pgfqpoint{0pt}{2pt}}
    \pgfpathlineto{\pgfqpoint{2pt}{0pt}}
    \pgfusepath{stroke}
  }

\pgfdeclarepatternformonly{myCrosshatch}
  {\pgfqpoint{-1pt}{-1pt}}
  {\pgfqpoint{2pt}{2pt}}
  {\pgfqpoint{1.8pt}{1.8pt}}
  {
    \pgfsetlinewidth{0.5pt}
    \pgfpathmoveto{\pgfqpoint{0pt}{0pt}}
    \pgfpathlineto{\pgfqpoint{2pt}{2pt}}
    \pgfusepath{stroke}
    \pgfpathmoveto{\pgfqpoint{0pt}{2pt}}
    \pgfpathlineto{\pgfqpoint{2pt}{0pt}}
    \pgfusepath{stroke}
  }

\pgfdeclarepatternformonly{myHorizontalLines}
  {\pgfqpoint{-1pt}{-1pt}}
  {\pgfqpoint{2pt}{2pt}}
  {\pgfqpoint{1.8pt}{1.8pt}}
  {
    \pgfsetlinewidth{1pt}
    \pgfpathmoveto{\pgfqpoint{0pt}{1pt}}
    \pgfpathlineto{\pgfqpoint{2pt}{1pt}}
    \pgfusepath{stroke}
  }

  \pgfdeclarepatternformonly{myDots}
  {\pgfqpoint{-1pt}{-1pt}}
  {\pgfqpoint{2pt}{2pt}}
  {\pgfqpoint{1.8pt}{1.8pt}}
  {
    \pgfsetlinewidth{0.5pt}
    \pgfpathcircle{\pgfqpoint{1pt}{1pt}}{0.5pt}
    \pgfusepath{fill}
  }

\usepackage{pifont}

\begin{document}
\title{\MYLOGNAME{}: Efficient Log Anomaly Detection Directly on Compressed Representations}

\author{
  Benzhao Tang, Shiyu Yang
}

\affiliation{
    Guangzhou University;
}
\affiliation{
    benzhaotang@outlook.com; syyang@gzhu.edu.cn;
  }

\begin{abstract}
The explosive growth of system logs makes streaming compression essential, yet existing log anomaly detection (LAD) methods incur severe pre-processing overhead by requiring full decompression and parsing.
We introduce \textbf{\MYLOGNAME{}}, the first deep learning framework to perform LAD \emph{directly} on compressed byte streams.
\MYLOGNAME{} bypasses these bottlenecks by exploiting a key insight: normal logs compress into regular byte patterns, while anomalies systematically disrupt them.
To extract these multi-scale deviations from opaque bytes, we propose a purpose-built architecture integrating a dilated convolutional byte encoder, a hybrid Transformer--mLSTM, and four-way aggregation pooling.
This is coupled with a two-stage training strategy of masked pre-training and focal-contrastive fine-tuning to effectively handle severe class imbalance.
Evaluated across five datasets, \MYLOGNAME{} achieves a state-of-the-art average F1-score of 0.9909 and outperforms the best baseline by 2.72 percentage points.
It delivers superior accuracy while completely eliminating decompression and parsing overheads, offering a robust solution that generalizes to structured streaming compressors.
\end{abstract}

\maketitle

\renewcommand{\vldbauthors}{Benzhao Tang, Shiyu Yang}
\pagestyle{\vldbpagestyle}
\begingroup\small\noindent\raggedright\textbf{PVLDB Reference Format:}\\
\vldbauthors. \vldbtitle. PVLDB, \vldbvolume(\vldbissue): \vldbpages, \vldbyear.\\
\href{https://doi.org/\vldbdoi}{doi:\vldbdoi}
\endgroup
\begingroup
\renewcommand\thefootnote{}\footnote{\noindent
$^{\ast}$Shiyu Yang is the corresponding author.\\
This work is licensed under the Creative Commons BY-NC-ND 4.0 International License. Visit \url{https://creativecommons.org/licenses/by-nc-nd/4.0/} to view a copy of this license. For any use beyond those covered by this license, obtain permission by emailing \href{mailto:info@vldb.org}{info@vldb.org}. Copyright is held by the owner/author(s). Publication rights licensed to the VLDB Endowment. \\
\raggedright Proceedings of the VLDB Endowment, Vol. \vldbvolume, No. \vldbissue\ %
ISSN 2150-8097. \\
\href{https://doi.org/\vldbdoi}{doi:\vldbdoi} \\
}\addtocounter{footnote}{-1}\endgroup

\ifdefempty{\vldbavailabilityurl}{}{
\vspace{.3cm}
\begingroup\small\noindent\raggedright\textbf{PVLDB Artifact Availability:}\\
The source code, data, and/or other artifacts have been made available at \url{\vldbavailabilityurl}.
\endgroup
}

\section{Introduction}\label{sec:intro}

System logs are indispensable for monitoring, diagnosing, and safeguarding modern software infrastructure.
Every layer of a distributed system---from application servers and container orchestrators to storage engines and network devices---continuously emits log entries that capture state transitions, error conditions, and security-relevant events.
Driven by the proliferation of microservice architectures and IoT deployments, the daily volume of log generation has surged to tens of petabytes: Uber produces over 10\,PB during peak periods~\cite{wang2024muslope}, and WeChat generates 16--20\,PB daily~\cite{yu2023logreducer}.
Given regulatory mandates requiring lossless retention for months to years~\cite{liu2019logzip,rodrigues2021clp,wei2023loggrep,wei2021logreducer}, the cost of collecting, transmitting, and archiving log data has become a first-order operational concern.

Streaming log compressors such as LogLite~\cite{tang2025loglite} have emerged as an effective response.
Operating line-by-line at the point of generation, these compressors produce compact byte streams that can be transmitted to cloud servers immediately, reducing bandwidth and storage costs by an order of magnitude without sacrificing log fidelity.
Meanwhile, log anomaly detection (LAD) remains a critical online analysis task: operators need to identify anomalous system behaviors---hardware faults, service degradation, security incidents---from the incoming log stream with minimal latency.

However, existing LAD methods impose a fundamental conflict with the compressed data path.
Whether they rely on parsed log templates~\cite{du2017deeplog,zhang2019robust,xie2022loggd} or operate directly on raw log text~\cite{le2021log,guo2021logbert}, all prior approaches require access to fully decompressed log messages.
This forces a \emph{decompress--parse--extract--detect} pipeline that partially negates the efficiency gains achieved on the data path.
The decompression and parsing stages alone consume a significant portion of the end-to-end detection time, constituting a throughput bottleneck in high-volume streaming environments.

A natural question arises: \emph{\textbf{is decompression truly necessary for anomaly detection?}}
Streaming compressors exploit the regularity of normal log data---fixed templates, limited variable ranges, and temporal locality---to achieve high compression ratios.
Anomalous entries, which by definition deviate from these regularities, produce compressed representations that \emph{systematically differ} from normal ones: novel keywords generate longer literal runs, unexpected variable values disrupt run-length patterns, and new templates yield entirely uncompressed output.
These structured deviations suggest that the compressed byte stream itself carries sufficient signal for anomaly detection---without the need for decompression or parsing.

Building on this insight, we propose \textbf{\MYLOGNAME{}} (\underline{C}ompressed \underline{L}og \underline{A}nomaly \underline{D}etection), the first deep learning framework that performs anomaly detection \emph{directly} on compressed log byte streams.
\MYLOGNAME{} takes the variable-length byte sequence emitted by a streaming compressor for a window of consecutive log entries, and outputs a binary anomaly prediction---bypassing the entire text-level processing pipeline.
Its architecture is purpose-built for the multi-scale structural regularity of compressor output, combining a dilated convolutional encoder, a hybrid Transformer--mLSTM sequential encoder, and a four-way aggregation pooling mechanism.
A two-stage training strategy---self-supervised pre-training via masked feature prediction followed by joint focal-contrastive fine-tuning---addresses the unique challenges of learning from semantically opaque bytes under severe class imbalance.
Although our primary evaluation uses LogLite-B as the streaming compressor, the architecture generalizes to any compressor whose output exhibits structured byte-level patterns---a property shared by the broad family of LZ-, RLE-, and dictionary-based compressors.

Compressed-domain computing has been explored in database query processing~\cite{zhou2024f,zhang2022compressdb,guan2023homomorphic} and log retrieval~\cite{wei2023loggrep,rodrigues2021clp,wang2024muslope}, but these systems are restricted to deterministic operations such as keyword matching and statistical aggregation.
\MYLOGNAME{} bridges the gap to deep analytical tasks, elevating compressed-domain computing from shallow retrieval to semantic-level anomaly detection.

\smallskip
\noindent\textbf{Contributions.}

\begin{itemize}[leftmargin=*]

\item We establish, for the first time, a direct path from compressed log ingestion to anomaly prediction, entirely eliminating the decompression and parsing stages required by all prior LAD methods and reducing end-to-end latency in high-throughput streaming environments.

\item We propose \MYLOGNAME{}, a five-stage neural architecture comprising a dilated convolutional byte encoder, a hybrid Transformer--mLSTM sequential encoder, and four-way aggregation pooling, purpose-built to capture the multi-scale structural patterns present in compressed byte streams.

\item We design a two-stage training strategy combining masked feature prediction with InfoNCE contrastive pre-training and joint focal-contrastive fine-tuning, together with contextual priority sampling and span masking augmentation, to address the challenges of learning from semantically opaque bytes under severe class imbalance.

\item On five widely used datasets, \MYLOGNAME{} achieves the highest F1-score on every dataset (average 0.9909), outperforming the best baseline by 2.72 percentage points---despite operating on compressed byte streams that have never been decompressed or parsed.

\end{itemize}
\section{\MYLOGNAME{} Model Architecture}\label{sec:architecture}

This section details the architecture of \MYLOGNAME{}, following the data flow through five stages: architectural overview (\S\ref{sec:overview}), byte embedding (\S\ref{sec:embedding}), multi-scale dilated CNN encoder (\S\ref{sec:cnn}), hybrid Transformer--mLSTM sequential encoder (\S\ref{sec:encoder}), four-way aggregation pooling (\S\ref{sec:pooling}), and the anomaly detection head (\S\ref{sec:head}).

\subsection{Architectural Overview}\label{sec:overview}

\MYLOGNAME{} operates on the compressed byte sequence corresponding to a \emph{window} of \(W\) consecutive log entries (by default \(W = 100\)).
A streaming compressor processes the window and emits a variable-length byte sequence \(\mathbf{b} = (b_1, b_2, \dots, b_L)\), where each element \(b_i \in \{0, 1, \dots, 255\}\).
\MYLOGNAME{} takes this byte sequence as input and produces a binary prediction indicating whether the window contains anomalous log entries.

The end-to-end pipeline proceeds through five stages.

\textbf{(i)~Byte Embedding.}
Each byte is mapped to a continuous vector via a learnable embedding table; a special classification token \texttt{[CLS]} is prepended to the sequence.

\textbf{(ii)~Multi-Scale Dilated CNN Encoder.}
The embedded byte sequence (excluding \texttt{[CLS]}) is compressed by a stack of dilated convolutional blocks that reduce the sequence length by \(16\times\) while extracting local byte-level patterns at multiple scales.

\textbf{(iii)~Hybrid Transformer--mLSTM Sequential Encoder.}
The projected \texttt{[CLS]} embedding is concatenated with the CNN output, augmented with positional encodings, and processed by a two-layer encoder comprising one Transformer self-attention layer and one mLSTM memory layer.

\textbf{(iv)~Four-Way Aggregation Pooling.}
The encoder output is aggregated through four complementary pooling mechanisms---CLS, learned attention, max, and mean---that capture different aspects of anomaly manifestation.

\textbf{(v)~Anomaly Detection Head.}
A lightweight linear classifier with multi-scale dropout produces the final prediction.

Formally, given compressed byte sequence \(\mathbf{b}\), \MYLOGNAME{} computes:
\begin{equation}\label{eq:pipeline}
  \hat{y}
  = f_{\mathrm{head}}\!\Bigl(
      f_{\mathrm{pool}}\!\bigl(
        f_{\mathrm{enc}}\!\bigl(
          f_{\mathrm{cnn}}\!\bigl(
            f_{\mathrm{emb}}(\mathbf{b})
          \bigr)
        \bigr)
      \bigr)
    \Bigr),
\end{equation}
where \(\hat{y} \in \{0, 1\}\) denotes the predicted anomaly label.


\subsection{Byte Embedding Layer}\label{sec:embedding}

The first stage converts discrete byte values into continuous representations amenable to neural network processing.
We maintain a learnable embedding matrix \(\mathbf{E} \in \mathbb{R}^{V \times d_e}\), where the vocabulary size \(V = 259\) covers four disjoint tokens: the 256 standard byte values (IDs 0--255) representing all possible byte outputs of the compressor, a padding token \texttt{[PAD]} (ID 256) for batching sequences of variable length, a classification token \texttt{[CLS]} (ID 257) serving as a global sequence-level anchor, and a mask token \texttt{[MASK]} (ID 258) used exclusively during self-supervised pre-training (\S\ref{sec:pretrain}).
The byte embedding dimension is set to \(d_e = 128\).

The input sequence is constructed by prepending \texttt{[CLS]} to the raw compressed byte sequence.
Sequences exceeding the maximum length \(L_{\max} = 8{,}192\) are truncated; shorter sequences are right-padded with \texttt{[PAD]}.
A scalar field \texttt{lengths} records the number of valid (non-padding) positions in each sample and is propagated through all subsequent stages to ensure that padding tokens do not contribute to pooling or loss computation.

\textbf{Design rationale.}
Unlike natural language processing where each token carries explicit semantic meaning, individual bytes in a compressed stream are semantically opaque---their meaning depends on context (e.g., whether a given byte represents an RLE count, a preserved original character, or a header field).
A learnable embedding allows the model to discover context-dependent byte representations during training.
The moderate embedding dimension (\(d_e = 128\), expanded to \(d = 512\) by the CNN) balances representational capacity against the risk of overfitting on the relatively small byte vocabulary.

\subsection{Multi-Scale Dilated CNN Encoder}\label{sec:cnn}

After embedding, the \texttt{[CLS]} token is separated and projected from \(d_e\) to the model dimension \(d = 512\) through a linear layer.
The remaining byte embeddings \(\mathbf{X} \in \mathbb{R}^{(L_{\max}-1) \times d_e}\) are transposed to channel-first layout and passed through a three-block convolutional encoder.

\subsubsection{\textbf{Multi-Scale Structure of Compressed Byte Streams}}

Anomalies disrupt compression patterns at multiple granularities.
The compressed byte stream mirrors this multi-scale structure.
At the \emph{finest} granularity (2--5 bytes), individual structural elements are encoded: 1-byte headers, single-byte RLE counts, and individual preserved characters.
At a \emph{medium} scale (10--30 bytes), complete compressed log entries emerge, each consisting of a header, an instruction bitmap, and a data payload.
At the \emph{coarsest} scale (50--100+ bytes), patterns spanning multiple entries reflect the overall composition of the window---the distribution of compressed vs.\ uncompressed entries, the frequency of different \texttt{Window\_ID} values, and the statistical profile of RLE run lengths.

Anomalies manifest at all three scales: a novel error keyword produces an uncompressed entry with distinctive fine-grained byte patterns; a malformed entry with unexpected variables disrupts the medium-scale RLE structure; and a burst of anomalous entries alters the coarse-scale statistical profile.

\subsubsection{\textbf{Architecture}}

To capture these multi-scale patterns efficiently, the CNN encoder employs three convolutional blocks with \emph{increasing dilation rates}, each consisting of a one-dimensional convolution, group normalization~\cite{wu2018group} (with a single group, equivalent to layer normalization along the channel dimension), and ReLU activation.

\textbf{Block~1 (fine-grained, dilation = 1).}
Conv1d with 256 output channels, kernel size 5, stride 2, and dilation 1.
The receptive field spans 5 byte positions, covering individual header fields, RLE counts, and preserved characters---the atomic units of the compressed representation.

\textbf{Block~2 (medium-range, dilation = 2).}
Conv1d with 512 output channels, kernel size 5, stride 2, and dilation 2.
The effective receptive field expands to approximately 18 original byte positions, sufficient to capture the complete structure of a single compressed log entry.

\textbf{Block~3 (coarse-scale, dilation = 4).}
Conv1d with 512 output channels, kernel size 5, stride 4, and dilation 4.
The effective receptive field reaches approximately 68 original byte positions, spanning multiple compressed entries and enabling cross-entry pattern recognition.

The cumulative stride of \(2 \times 2 \times 4 = 16\) reduces the sequence length from approximately 8,191 positions to \(T' \approx 512\), transforming a raw byte sequence that would require \(O(L^2)\) self-attention computation into a manageable length.
Concurrently, the increasing dilation rates \((1, 2, 4)\) expand the receptive field to well over 100 positions \emph{without} increasing the parameter count.
This \(16\times\) compression is a critical enabler for the subsequent Transformer layer, whose quadratic attention cost on the original 8,192-length sequence would be prohibitive.

\subsection{Hybrid Transformer--mLSTM Sequential Encoder}\label{sec:encoder}

The projected \texttt{[CLS]} embedding is concatenated with the CNN output to form the encoder input:
\begin{equation}
  \mathbf{H}^{(0)}
  = \bigl[\,\mathbf{c}_{\texttt{cls}};\; \mathbf{s}_1, \dots,
    \mathbf{s}_{T'}\,\bigr]
  \in \mathbb{R}^{(1+T') \times d},
\end{equation}
augmented with sinusoidal positional encodings~\cite{vaswani2017attention}.
The encoder is a stack of two heterogeneous layers: one Transformer self-attention layer followed by one mLSTM memory layer.

\subsubsection{\textbf{Transformer Self-Attention Layer}}\label{sec:transformer_layer}

The Transformer layer adopts a Pre-Norm architecture~\cite{xiong2020layer} with RMSNorm~\cite{zhang2019root} for stable training and a SwiGLU~\cite{shazeer2020glu} feed-forward network for improved expressivity.
Multi-head self-attention uses \(H = 8\) heads with model dimension \(d = 512\), and the SwiGLU FFN has inner dimension \(d_{\mathrm{ff}} = 2{,}048\).
The forward pass is:
\begin{align}
  \mathbf{H}' &= \mathbf{H} + \mathrm{Dropout}\!\bigl(
    \mathrm{MHA}\!\bigl(\mathrm{RMSNorm}(\mathbf{H})\bigr)
  \bigr), \label{eq:tf_attn} \\[4pt]
  \mathbf{H}^{(1)} &= \mathbf{H}' + \mathrm{SwiGLU}\!\bigl(
    \mathrm{RMSNorm}(\mathbf{H}')
  \bigr). \label{eq:tf_ffn}
\end{align}

The Transformer layer captures \textbf{long-range pairwise interactions} through full quadratic attention.
In compressed log streams, certain anomalies are detectable only through cross-position reasoning---for example, relating a compressed error code in one entry to a compressed configuration value in a distant entry.
The self-attention mechanism excels at modeling such dependencies, and the relatively short input length (\(T' \approx 512\) after CNN compression) makes the \(O(T'^2)\) cost manageable.

\subsubsection{\textbf{mLSTM Memory Layer}}\label{sec:mlstm_layer}

The mLSTM layer~\cite{beck2024xlstm} complements the Transformer with \(O(T')\) linear-time processing through a matrix-valued associative memory.
Before the attention computation, a depthwise separable convolution with kernel size 3 injects local context, which is important for resolving sub-token byte groupings that arise from multi-byte RLE counts and instruction-data boundaries in the compressed stream.

The core mechanism uses squared ReLU kernels for query and key projections:
\begin{equation}\label{eq:squared_relu}
  \hat{\mathbf{q}}_t
  = \bigl[\mathrm{ReLU}(\mathbf{q}_t)\bigr]^2,
  \qquad
  \hat{\mathbf{k}}_t
  = \bigl[\mathrm{ReLU}(\mathbf{k}_t)\bigr]^2,
\end{equation}
and computes the output via a matrix memory:
\begin{equation}\label{eq:memory}
  \mathbf{C} = \hat{\mathbf{K}}^\top \mathbf{V},
  \qquad
  \mathbf{z} = \textstyle\sum_t \hat{\mathbf{k}}_t,
  \qquad
  \mathbf{o}_t
  = \frac{\hat{\mathbf{q}}_t^\top \mathbf{C}}
         {\hat{\mathbf{q}}_t^\top \mathbf{z} + \epsilon},
\end{equation}
where \(\mathbf{C} \in \mathbb{R}^{d_h \times d_h}\) accumulates a compressed summary of all key--value associations.
The mLSTM layer uses the same Pre-Norm residual structure and SwiGLU FFN as the Transformer layer.

\subsubsection{\textbf{Rationale for the Hybrid Design}}\label{sec:hybrid_rationale}

The two layers serve complementary roles.
The Transformer's full attention computes \emph{explicit pairwise} relationships between all CNN output positions, essential for detecting anomalies that manifest as unusual correlations between specific byte patterns at distant positions.
The mLSTM's linear attention with matrix memory efficiently captures \emph{global distributional statistics}---the overall distribution of byte values, the frequency of different RLE run lengths, and the proportion of compressed vs.\ uncompressed entries---which reflect the macro-level health of the log window.
These statistics are naturally summarized by the additive key--value accumulation in Equation~\eqref{eq:memory}.

By stacking one layer of each type rather than multiple layers of a single type, \MYLOGNAME{} achieves an effective balance between representational power and computational efficiency, keeping the model lightweight enough for online deployment where the compressed stream must be analyzed in real time.

\subsection{Four-Way Aggregation Pooling}\label{sec:pooling}

Anomalies in compressed log streams manifest in fundamentally different ways: some produce sharp localized byte-pattern deviations, others cause diffuse distributional shifts across the entire window, and still others are detectable only through selective focus on specific positions.
A single pooling strategy inevitably introduces bias toward one manifestation type.
We propose a \textbf{four-way aggregation pooling} mechanism that extracts complementary views of the encoded sequence \(\mathbf{H} \in \mathbb{R}^{(1+T') \times d}\).

\textbf{Path~1: CLS pooling.}
The \texttt{[CLS]} output \(\mathbf{h}_{\texttt{cls}} = \mathbf{H}[0]\) provides a holistic summary.
Having attended to all positions through both encoder layers, it integrates fine-grained pairwise interactions with global distributional statistics.

\textbf{Path~2: Learned attention pooling.}
A two-layer scoring network computes position-wise importance weights:
\begin{equation}\label{eq:attn_pool}
  \alpha_t
  = \frac
    {\exp\!\bigl(\mathbf{w}_2^\top \tanh(\mathbf{W}_1
      \mathbf{h}_t)\bigr)}
    {\sum_{t'} \exp\!\bigl(\mathbf{w}_2^\top \tanh(\mathbf{W}_1
      \mathbf{h}_{t'})\bigr)},
  \qquad
  \mathbf{a} = \textstyle\sum_t \alpha_t\, \mathbf{h}_t,
\end{equation}
where \(\mathbf{W}_1 \in \mathbb{R}^{(d/4) \times d}\) and \(\mathbf{w}_2 \in \mathbb{R}^{d/4}\).
This path learns to selectively focus on positions where anomalous byte patterns deviate from the expected compressed structure.

\textbf{Path~3: Max pooling.}
Element-wise maximum over valid positions captures the most salient activation per feature dimension, effective at detecting \emph{spike} anomalies such as previously unseen error messages emitted as uncompressed literal content.

\textbf{Path~4: Mean pooling.}
Length-normalized average over valid positions captures the aggregate distributional profile, effective for detecting \emph{diffuse} anomalies where no single entry is dramatically anomalous but the overall window deviates from the norm.

The four representations are concatenated:
\begin{equation}\label{eq:pool_concat}
  \mathbf{p}
  = [\,\mathbf{h}_{\texttt{cls}};\; \mathbf{a};\; \mathbf{m};\;
    \boldsymbol{\mu}\,]
  \in \mathbb{R}^{4d}.
\end{equation}

\subsection{Anomaly Detection Head with Multi-Scale Dropout}\label{sec:head}

The detection head maps the \(4d\)-dimensional pooled vector to a binary anomaly prediction via a single linear layer:
\begin{equation}\label{eq:head}
  \mathbf{z} = \mathbf{W}_{\mathrm{cls}}\,\mathbf{p}
  + \mathbf{b}_{\mathrm{cls}},
  \qquad \mathbf{W}_{\mathrm{cls}} \in \mathbb{R}^{2 \times 4d}.
\end{equation}

\textbf{Multi-Scale Dropout.}
During training, instead of a single forward pass with one dropout mask, we perform \(K = 5\) independent forward passes through the linear layer, each with a different random dropout mask applied to \(\mathbf{p}\) (dropout rate \(p = 0.15\)), and average the resulting logits:
\begin{equation}\label{eq:ms_dropout}
  \mathbf{z}_{\mathrm{train}}
  = \frac{1}{K} \sum_{k=1}^{K}
    \bigl(\mathbf{W}_{\mathrm{cls}} \cdot
    \mathrm{Dropout}_k(\mathbf{p})
    + \mathbf{b}_{\mathrm{cls}}\bigr).
\end{equation}
This functions as an \emph{implicit ensemble} at the feature level: each dropout mask forces the classifier to rely on a different subset of the \(4d\) features from the four pooling paths, preventing co-adaptation between pooling signals and producing more robust, better-calibrated predictions.
The mechanism is particularly important for anomaly detection, where the cost of false negatives is typically high: ensemble averaging reduces logit variance, thereby reducing the likelihood that a borderline anomalous window is misclassified.
During inference, dropout is disabled and a single forward pass is performed, incurring no additional cost.

\section{Training Strategy and Optimization}\label{sec:training}

Training \MYLOGNAME{} on compressed log byte streams presents unique challenges.
First, individual bytes lack the discrete semantic grounding of word tokens, making standard masked language modeling objectives ill-suited.
Second, log anomaly detection datasets exhibit severe class imbalance, with anomalous windows often constituting less than 5\% of the data.
Third, anomalous events cluster temporally, creating a non-uniform distribution that standard random sampling ignores.

We address these challenges with a two-stage pipeline: self-supervised pre-training that learns general-purpose byte-level representations (\S\ref{sec:pretrain}), followed by fine-tuning with a joint focal-contrastive objective (\S\ref{sec:finetune}).
We additionally introduce contextual priority sampling (\S\ref{sec:sampling}), span masking augmentation (\S\ref{sec:augmentation}), and a comprehensive optimization procedure (\S\ref{sec:optimization}).

\subsection{Self-Supervised Pre-Training via Masked Feature Prediction}\label{sec:pretrain}

The goal of pre-training is to learn byte-level representations that capture the structural regularities of compressed log streams, providing a strong initialization for supervised fine-tuning.

\subsubsection{\textbf{Why Not Masked Byte Modeling?}}

A natural starting point is masked language modeling adapted to the byte level: mask random positions and predict the original byte value.
This approach is poorly suited for compressed streams for two reasons.
First, the 256-class prediction task is dominated by a few high-frequency values (e.g., the null byte \texttt{0x00} from RLE, common ASCII characters from XOR-Preserve), and cross-entropy loss provides weak gradients for rare but informative bytes.
Second, predicting the exact byte value requires modeling low-level encoding details (e.g., precise RLE counts) that are largely irrelevant for anomaly detection; what matters is whether the \emph{pattern} of a compressed region is normal, not its exact byte values.

\subsubsection{\textbf{Masked Feature Prediction}}

We instead design a \textbf{masked feature prediction} task operating in the continuous feature space of the CNN encoder.
The input byte sequence is first embedded and passed through the CNN to obtain the downsampled feature sequence \(\mathbf{S} = (\mathbf{s}_1, \dots, \mathbf{s}_{T'}) \in \mathbb{R}^{T' \times d}\).
A proportion \(r = 15\%\) of valid positions are randomly selected and replaced with a learnable mask embedding \(\mathbf{e}_{\mathrm{mask}} \in \mathbb{R}^d\).
The masked sequence, together with the projected \texttt{[CLS]} token and positional encodings, is processed by the full hybrid encoder.
A linear prediction head \(\mathbf{W}_{\mathrm{pred}} \in \mathbb{R}^{d \times d}\) maps each encoder output at a masked position back to the \(d\)-dimensional feature space.

Masking at the \emph{CNN feature level} rather than the raw byte level is deliberate.
Each CNN output position aggregates information from a receptive field of dozens of bytes, capturing a local compressed pattern such as a complete RLE-encoded entry.
Predicting this feature vector requires the model to understand the structural role of the masked region within the window---precisely the representation needed for downstream anomaly detection.

\subsubsection{\textbf{InfoNCE Contrastive Loss}}

A na\"ive MSE objective is vulnerable to \textbf{representation collapse}: the model can minimize loss by mapping all positions to a constant vector.
This risk is acute for compressed streams, whose feature distribution is concentrated around common patterns.
We adopt the InfoNCE loss~\cite{oord2018representation}.
Let \(\mathbf{P} = \{\hat{\mathbf{s}}_1, \dots, \hat{\mathbf{s}}_N\}\) and \(\mathbf{T} = \{\mathbf{s}_1, \dots, \mathbf{s}_N\}\) denote the \(L_2\)-normalized predicted and original feature vectors at the \(N\) masked positions within a mini-batch.
The pre-training loss is:
\begin{equation}\label{eq:infonce}
  \mathcal{L}_{\mathrm{pre}}
  = -\frac{1}{N} \sum_{i=1}^{N}
    \log \frac{\exp(\hat{\mathbf{s}}_i^\top \mathbf{s}_i / \tau)}
              {\sum_{j=1}^{N} \exp(\hat{\mathbf{s}}_i^\top
                \mathbf{s}_j / \tau)},
\end{equation}
with temperature \(\tau = 0.1\).
If the model collapses to constant output, numerator and denominator terms become identical and the loss degrades to \(\log N\)---the theoretical maximum---generating a strong corrective gradient.
The contrastive objective simultaneously encourages discriminative features that distinguish different compressed patterns, directly benefiting the downstream separation of normal and anomalous representations.

\subsection{Fine-Tuning with Joint Focal-Contrastive Learning}\label{sec:finetune}

After pre-training, the entire model is fine-tuned end-to-end on labeled compressed log windows.

\subsubsection{\textbf{Joint Loss Function}}\label{sec:loss}

The fine-tuning objective combines a classification loss and a contrastive loss:
\begin{equation}\label{eq:joint_loss}
  \mathcal{L} = \mathcal{L}_{\mathrm{focal}}
  + \lambda(t) \cdot \mathcal{L}_{\mathrm{supcon}},
\end{equation}
where \(\lambda(t)\) is a time-dependent weight.

\textbf{Focal Loss with Label Smoothing.}
We adopt Focal Loss~\cite{lin2017focal} with focusing parameter \(\gamma = 2.0\) and label smoothing factor \(\epsilon_{\mathrm{ls}} = 0.05\).
In datasets where anomalous windows constitute as little as 2--3\% of the data (e.g., HDFS), standard cross-entropy is dominated by normal samples.
Focal Loss down-weights well-classified samples by \((1 - p_t)^\gamma\), directing gradient magnitude toward hard, misclassified anomalous samples.
Label smoothing prevents overconfident predictions near the decision threshold.

\textbf{Supervised Contrastive Loss.}
To learn a feature space where normal and anomalous representations form well-separated clusters, we apply Supervised Contrastive (SupCon) loss~\cite{khosla2020supervised}.
The pooled vector \(\mathbf{p}\) is passed through a two-layer projection head mapping to a 128-dimensional \(L_2\)-normalized embedding, on which the SupCon loss with temperature \(\tau = 0.07\) is computed.
This is particularly valuable because the boundary between normal and anomalous compressed representations is often subtle: an anomalous entry differing from its cached reference by only a few characters may produce a compressed representation that differs by just a handful of bytes.
The contrastive loss explicitly amplifies these differences.

\textbf{Dynamic contrastive weight decay.}
The contrastive weight \(\lambda(t)\) is annealed from \(\lambda_0 = 0.05\) to \(\lambda_{\min} = 0.005\) via a cosine schedule:
\begin{equation}\label{eq:lambda_decay}
  \lambda(t) = \lambda_{\min}
    + \tfrac{1}{2}\,(\lambda_0 - \lambda_{\min})\,
      \bigl(1 + \cos\tfrac{\pi t}{T}\bigr).
\end{equation}
Contrastive learning is most beneficial early in fine-tuning, when the feature space needs restructuring; as training progresses, the classification loss should dominate to refine the decision boundary.

\subsubsection{\textbf{Contextual Priority Sampling}}\label{sec:sampling}

Class imbalance in log anomaly detection exhibits a distinctive \textbf{temporal structure}: anomalous events occur in bursts, and windows adjacent to an anomalous window often contain early warning signals or residual effects.
We exploit this structure with \textbf{contextual priority sampling}.
At each epoch, a fraction \(\rho = 80\%\) of the training data is sampled non-uniformly: 40\% from a \emph{priority pool} \(\mathcal{P}\) consisting of all anomalous windows and their temporal neighbors within \(\pm 3\) positions,
\begin{equation}
  \mathcal{P} = \bigcup_{i \,:\, y_i = 1}
    \{j : |j - i| \leq 3\},
\end{equation}
and 60\% uniformly from \(\mathcal{D} \setminus \mathcal{P}\).
The remaining 20\% is not sampled, serving as an implicit regularizer.
This strategy is especially effective in the compressed domain, where near-anomaly windows often exhibit subtle shifts in RLE run length distributions that provide valuable training signal for early anomaly detection.

\subsubsection{\textbf{Data Augmentation via Span Masking}}\label{sec:augmentation}

During fine-tuning, \(r = 15\%\) of byte positions are replaced with \texttt{[MASK]} in contiguous spans of 2--5 bytes.
Unlike independent random masking, span masking removes entire \emph{compressed tokens}---an RLE count, a header, or a group of preserved characters---forcing the model to infer the masked region's contribution from context.
This prevents overfitting to superficial byte patterns and improves robustness to variations in compression output arising from changes in the compressor's sliding window state.

\subsubsection{\textbf{Optimization and Model Selection}}\label{sec:optimization}

We use AdamW~\cite{loshchilov2017decoupled} with learning rate \(3 \times 10^{-4}\), weight decay 0.01, \(\beta_1 = 0.9\), \(\beta_2 = 0.999\), and gradient clipping at norm 1.0.
The learning rate follows a 3-epoch linear warmup followed by cosine annealing to zero.

An Exponential Moving Average (EMA) of model parameters with decay \(\beta_{\mathrm{EMA}} = 0.998\) is maintained throughout training.
The EMA model is used for validation and testing, as it smooths stochastic gradient noise and produces more stable predictions.

\textbf{Model selection criterion.}
We use a composite score that jointly penalizes overfitting:
\begin{equation}\label{eq:score}
  \mathrm{Score}
  = F_{1}^{\mathrm{val}}
    - 0.2 \cdot \mathcal{L}^{\mathrm{val}}
    - 0.1 \cdot \mathcal{L}^{\mathrm{train}}.
\end{equation}
The inclusion of both loss terms discourages models that achieve high validation F1 through overfitting or lucky evaluation.
Early stopping triggers after 7 consecutive epochs without improvement.
\section{Experimental Evaluation}\label{ch:experiments}

We evaluate \MYLOGNAME{} through a comprehensive set of experiments: experimental setup (\S\ref{sec:setup}) and detection effectiveness (\S\ref{sec:effectiveness}).

\subsection{Experimental Setup}\label{sec:setup}

\subsubsection{\textbf{Hardware Environment}}

All experiments are conducted on a server with an Intel Xeon Gold 6133 CPU @ 2.50\,GHz, 93\,GB main memory, and an NVIDIA Tesla V100-SXM2 GPU (32\,GB).
All models are trained and evaluated under identical conditions.

\subsubsection{\textbf{Datasets and Preprocessing}}\label{sec:datasets}

We evaluate on five widely used public benchmarks spanning diverse system architectures, log volumes, and anomaly characteristics~\cite{Loghub2023,oliner2007supercomputers,xu2009online}.
Table~\ref{tab:datasets} summarizes key statistics.

\begin{table}[t]
  \centering
  \caption{Summary of the five benchmark datasets. \#Msg.\ denotes the total number of log entries; \#Seq.\ denotes the number of log sequences after windowing; \#Anom.\ denotes the number of anomalous sequences.}
  \label{tab:datasets}
  \renewcommand{\arraystretch}{1.15}
  \resizebox{\columnwidth}{!}{%
  \begin{tabular}{lrrrr}
    \toprule
    \textbf{Dataset} & \textbf{Size} & \textbf{\#Msg.} & \textbf{\#Seq.} & \textbf{\#Anom.} \\
    \midrule
    BGL         & 0.69\,GB & 4,747,963   & 47,479   & 4,802   \\
    Thunderbird & 1.40\,GB & 10,000,000  & 100,000  & 12,889  \\
    Liberty     & 1.11\,GB & 10,000,000  & 100,000  & 66,438  \\
    Spirit      & 1.37\,GB & 10,000,000  & 100,000  & 72,283  \\
    HDFS        & 1.47\,GB & 11,175,629  & 575,061  & 16,838  \\
    \bottomrule
  \end{tabular}%
  }
\end{table}

\textbf{BGL.}
The BlueGene/L dataset~\cite{oliner2007supercomputers} comprises 4,747,963 log messages from a supercomputer at Lawrence Livermore National Laboratory.
We apply a \emph{sliding window} strategy with window size 100, yielding 47,479 sequences (4,802 anomalous, 10.11\% anomaly ratio).

\textbf{Thunderbird, Liberty, and Spirit.}
These three supercomputer datasets each contain over 200 million entries~\cite{oliner2007supercomputers}.
To maintain computational tractability while preserving lifecycle coverage, we design a \emph{fixed-interval global sampling strategy}: after collecting a contiguous window of 100 entries, a fixed step is skipped before the next window.
This yields subsets of exactly 10,000,000 messages per dataset, from which we extract 100,000 sequences each.
Thunderbird contains 12,889 anomalous sequences (12.89\%), Liberty 66,438 (66.44\%), and Spirit 72,283 (72.28\%).
The high anomaly ratios of Liberty and Spirit provide a challenging testbed for evaluating model behavior under dense anomaly conditions.

\textbf{HDFS.}
The Hadoop Distributed File System dataset~\cite{xu2009online} contains 11,175,629 messages from a 203-node cluster.
We adopt a \emph{session window} strategy, grouping entries by \texttt{block\_id} to form 575,061 sequences (16,838 anomalous, 2.93\%).

For BGL, Thunderbird, Liberty, and Spirit, a sequence is labeled anomalous if it contains at least one anomalous entry; for HDFS, labels are assigned at the session level.
Data splitting follows a strict chronological 8:2 ratio for the supercomputer datasets (to prevent temporal leakage) and random 8:2 for HDFS.

After segmentation and splitting, each sequence is compressed by LogLite-B, which produces a variable-length compressed byte stream.
\MYLOGNAME{} consumes this byte stream directly as model input.
In contrast, all baselines operate on either decompressed raw text or parsed templates.

\subsubsection{\textbf{Baselines}}

We compare against three representative baselines covering the major paradigms in deep learning--based LAD.

\textbf{CNN~\cite{lu2018detecting}} operates on parsed log templates using a convolutional neural network, representing parsing-dependent CNN-based methods.

\textbf{LogRobust~\cite{zhang2019robust}} employs an attention-based Bi-LSTM with TF-IDF weighted semantic vectors of parsed templates, representing parsing-dependent RNN-based methods.

\textbf{NeuralLog~\cite{le2021log}} uses pre-trained BERT embeddings on raw log messages with a Transformer encoder, representing parsing-free methods that still require uncompressed text.

Among these, CNN and LogRobust require a log parser; NeuralLog and \MYLOGNAME{} do not.
Crucially, \MYLOGNAME{} is the only method that operates directly on \emph{compressed} byte streams.

\subsubsection{\textbf{Evaluation Metrics}}

We report Precision, Recall, and F1-score at the window level (or session level for HDFS):
\begin{equation}
  \text{Precision} = \frac{TP}{TP + FP},
\end{equation}

\begin{equation}
  \text{Recall} = \frac{TP}{TP + FN},
\end{equation}

\begin{equation}
  \text{F1} = \frac{2 \cdot \text{Precision} \cdot \text{Recall}}{\text{Precision} + \text{Recall}}.
\end{equation}
We additionally report the average F1-score across all five datasets.

\subsection{Detection Effectiveness}\label{sec:effectiveness}

Table~\ref{tab:main_results} presents the detection performance of all methods.

\begin{table*}[t]
  \centering
  \caption{Detection performance comparison across five benchmark datasets. ``Parser'' indicates whether the method requires a log parser. The best F1-score on each dataset is shown in \textbf{bold}. CLAD operates directly on compressed byte streams without decompression or parsing.}
  \label{tab:main_results}
  \renewcommand{\arraystretch}{1.15}
  \resizebox{\textwidth}{!}{%
  \begin{tabular}{lc ccc ccc ccc ccc ccc c}
    \toprule
    & & \multicolumn{3}{c}{\textbf{BGL}}
    & \multicolumn{3}{c}{\textbf{Thunderbird}}
    & \multicolumn{3}{c}{\textbf{Liberty}}
    & \multicolumn{3}{c}{\textbf{Spirit}}
    & \multicolumn{3}{c}{\textbf{HDFS}}
    & \textbf{Avg.} \\
    \cmidrule(lr){3-5} \cmidrule(lr){6-8} \cmidrule(lr){9-11}
    \cmidrule(lr){12-14} \cmidrule(lr){15-17} \cmidrule(lr){18-18}
    \textbf{Method} & \textbf{Parser}
    & Prec. & Rec. & F1
    & Prec. & Rec. & F1
    & Prec. & Rec. & F1
    & Prec. & Rec. & F1
    & Prec. & Rec. & F1
    & F1 \\
    \midrule
    CNN~\cite{lu2018detecting}
    & \ding{52}
    & 0.9530 & \textbf{0.9564} & 0.9547
    & 0.9986 & 0.9968 & 0.9977
    & 0.9478 & 0.9990 & 0.9728
    & 0.9837 & 0.9996 & 0.9916
    & 0.8932 & 0.9101 & 0.9016
    & 0.9637 \\
    LogRobust~\cite{zhang2019robust}
    & \ding{52}
    & 0.9261 & 0.9489 & 0.9373
    & 0.7385 & \textbf{0.9991} & 0.8492
    & 0.9463 & \textbf{0.9994} & 0.9721
    & 0.9988 & 0.9992 & 0.9990
    & 0.9242 & 0.9541 & 0.9389
    & 0.9393 \\
    NeuralLog~\cite{le2021log}
    & \ding{56}
    & 0.9022 & 0.8250 & 0.8619
    & 0.9994 & 0.9658 & 0.9823
    & 0.9957 & 0.9595 & 0.9772
    & 0.9874 & 0.9996 & 0.9934
    & 0.9901 & 0.9967 & 0.9933
    & 0.9616 \\
    \midrule
    \textbf{CLAD (Ours)}
    & \ding{56}
    & \textbf{0.9768} & 0.9530 & \textbf{0.9645}
    & \textbf{0.9997} & 0.9986 & \textbf{0.9991}
    & \textbf{0.9970} & 0.9971 & \textbf{0.9970}
    & \textbf{0.9996} & \textbf{0.9999} & \textbf{0.9998}
    & \textbf{0.9906} & \textbf{0.9974} & \textbf{0.9940}
    & \textbf{0.9909} \\
    \bottomrule
  \end{tabular}%
  }
\end{table*}

\textbf{Overall comparison.}
\MYLOGNAME{} achieves the highest F1-score on all five datasets and the highest average F1 of 0.9909, surpassing the strongest baseline (CNN, 0.9637) by 2.72 percentage points and the best parsing-free baseline (NeuralLog, 0.9616) by 2.93 points.
This is noteworthy because \MYLOGNAME{} operates on compressed byte streams that have never been decompressed or parsed, yet consistently outperforms methods with access to full uncompressed text or structured parsed templates.
The result validates our central hypothesis: the compressed representation preserves---and concentrates---the structural and semantic signals necessary for anomaly detection.

\textbf{Supercomputer datasets (BGL, Thunderbird, Liberty, Spirit).}
\MYLOGNAME{} achieves F1-scores of 0.9645, 0.9991, 0.9970, and 0.9998, respectively.
The most significant improvements appear on BGL and Liberty.
On BGL, \MYLOGNAME{} outperforms the second-best method (CNN) by 0.98 points in F1, driven by a 2.38-point precision improvement.
BGL contains diverse anomaly types---hardware faults, memory errors, network failures---that produce distinct signatures in the compressed domain.
The four-way aggregation pooling is specifically designed for this heterogeneity: hardware faults create sharp byte-level spikes (captured by max pooling), while gradual degradation shifts the distributional profile (captured by mean pooling).

On Liberty, \MYLOGNAME{} improves upon CNN by 2.42 points, primarily through a dramatic precision increase (0.9970 vs.\ 0.9478) while maintaining near-perfect recall (0.9971), indicating far fewer false positives---an essential property for operational deployment.
On Thunderbird and Spirit, all methods achieve high F1, but \MYLOGNAME{} still attains the best results with near-perfect precision and recall.

\textbf{Distributed system dataset (HDFS).}
HDFS represents a fundamentally different architecture (distributed storage) with session-level windowing.
\MYLOGNAME{} achieves 0.9940, slightly surpassing NeuralLog (0.9933) and substantially outperforming CNN (0.9016) and LogRobust (0.9389).
The strong performance confirms that \MYLOGNAME{} generalizes across system types and windowing strategies.

\textbf{\MYLOGNAME{} vs.\ parsing-dependent methods.}
Despite having access to clean parsed templates, both CNN and LogRobust are outperformed on every dataset.
The most dramatic case is LogRobust on Thunderbird, where precision drops to 0.7385 due to template instability---a failure mode that \MYLOGNAME{}, which bypasses parsing entirely, is immune to.

\textbf{\MYLOGNAME{} vs.\ the parsing-free baseline (NeuralLog).}
\MYLOGNAME{} outperforms NeuralLog on all datasets, with the largest margin on BGL (0.9645 vs.\ 0.8619, a gap of 10.26 points).
NeuralLog's low recall on BGL (0.8250) suggests that word-level semantics struggle with diverse supercomputer vocabulary.
In contrast, \MYLOGNAME{} operates at the byte level where the compressor has already normalized lexical variability into compact patterns---a natural feature extraction that is robust to vocabulary drift.

\textbf{Precision--recall balance.}
Across all datasets, \MYLOGNAME{} maintains precision above 0.976 and recall above 0.953, without the trade-off that affects several baselines (e.g., CNN achieves high recall on Liberty but low precision; NeuralLog achieves high precision on HDFS but low recall on BGL).
This balanced performance is attributable to the joint focal-contrastive training: Focal Loss preserves recall by attending to the minority class, while SupCon loss maintains precision by enforcing clear separation between normal and anomalous representations.
\section{Related Work}\label{ch:related}

We review two lines of research most relevant to \MYLOGNAME{}: log anomaly detection and compressed-domain data analysis.

\subsection{Log Anomaly Detection}

Existing methods can be broadly categorized into sequence-based, graph-based, and LLM-enhanced approaches.

\textbf{Sequence-based methods} constitute the majority of LAD research.
A typical pipeline parses raw messages into event templates and applies detection models to the resulting sequences.
DeepLog~\cite{du2017deeplog} pioneered this paradigm with LSTM-based next-event prediction.
LogAnomaly~\cite{meng2019loganomaly} augmented it by jointly modeling sequential and quantitative patterns with semantic embeddings.
LogRobust~\cite{zhang2019robust} addressed log instability via attention-based Bi-LSTM with TF-IDF weighted vectors.
PLELog~\cite{yang2021semi} and Pluto~\cite{ma2024pluto} tackled semi-supervised and noisy-label settings, respectively, while RT-Log~\cite{jia2023robust} and LogTransfer~\cite{chen2020logtransfer} explored cross-system transferability.
MultiLog~\cite{zhang2024multivariate} extended detection to distributed environments through multi-node aggregation.

To decouple detection from error-prone parsing, parsing-free approaches have emerged.
NeuralLog~\cite{le2021log} embeds raw messages via pre-trained word representations and applies a Transformer encoder.
LogBERT~\cite{guo2021logbert} pre-trains a BERT model via masked log event prediction.
Although these methods eliminate explicit parsing, they still require the full-text, decompressed log messages as input.

\textbf{Graph-based methods} capture structural dependencies among log events.
LogGD~\cite{xie2022loggd} employs a Graph Transformer for graph-level classification.
Glad-PAW~\cite{wan2021glad} integrates positional information via weighted graph attention.
TP-GNN~\cite{liu2024tp} incorporates temporal dynamics, and SLAD~\cite{tang2024substructure} discovers representative substructures via Monte Carlo Tree Search for fine-grained detection.
These methods still rely on parsed or raw text to construct the underlying graphs.

\textbf{LLM-enhanced methods} leverage large language models for richer semantic understanding.
LogGPT~\cite{qi2023loggpt} explored in-context learning for anomaly reasoning.
LogSynergy~\cite{sui2025bridging} standardizes log syntax across systems via LLM-based event interpretation.
CoLA~\cite{zhu2025cola} combines a LogMoE filter with a domain-specialized LAD-LLM.
Despite strong accuracy and explainability, the prohibitive inference cost of LLMs limits their applicability to high-volume real-time streams.

All aforementioned methods---whether sequence-based, graph-based, or LLM-enhanced, parsing-dependent or parsing-free---share a fundamental requirement: \textit{decompressed, raw-text log messages as input}.
\MYLOGNAME{} is the first framework to eliminate the entire text-level processing pipeline, establishing a direct path from compressed log ingestion to anomaly prediction.

\subsection{Compressed-Domain Data Analysis}
\label{subsec:compressed_domain}

Compressed-domain computing spans three primary directions.

\textbf{Grammar-based database computing.}
TADOC~\cite{zhou2024f} pioneered converting text into hierarchical grammar rules (DAGs) to support operations like frequency counting and string matching via graph traversal.
CompressDB~\cite{zhang2022compressdb} extended this to CRUD operations through a ``data hole'' mechanism.
DPTC~\cite{hu2025cost} introduced decompression-transparent compression for fine-grained row-level random access.
Most recently, HOCO~\cite{guan2023homomorphic} formalized homomorphic compression with properties such as \textit{Directness} and \textit{Strong Homomorphism}, achieving substantial throughput improvements via hash offset mapping and late materialization.

\textbf{Log-oriented retrieval.}
LogGrep~\cite{wei2023loggrep} proposed capsule-based indexing for feature matching directly at the compressed level.
CLP~\cite{rodrigues2021clp} designed a column-oriented log archive supporting SQL-like queries on compressed data.
LogCloud~\cite{wang2025logcioud} introduced FM-index--based storage enabling keyword search on object storage without decompression.

\textbf{Hardware acceleration.}
F-TADOC~\cite{zhou2024f} provides an FPGA-based framework for high-throughput CFG traversal.
IFGC~\cite{fan2025inference} explored inference-friendly graph compression for executing GNNs on compressed graph structures.

While these works have advanced database query acceleration and log retrieval, they are restricted to deterministic operations---keyword matching, regular expressions, and statistical aggregations.
They address \emph{where} relevant entries are in the compressed data but cannot answer \emph{what} the entries imply.
By introducing a semantic-aware feature extraction layer operating natively on compressed byte streams, \MYLOGNAME{} bridges this gap, elevating compressed-domain computing from shallow retrieval to deep analytical tasks such as anomaly detection.
\section{Conclusion}\label{sec:conclusion}

We introduce \MYLOGNAME{}, the first deep learning framework to perform log anomaly detection directly on compressed byte streams, entirely bypassing the costly decompression and parsing bottlenecks that plague conventional pipelines.
By leveraging a purpose-built architecture that features a multi-scale dilated CNN, a hybrid Transformer-mLSTM, and a two-stage pre-training and fine-tuning strategy, \MYLOGNAME{} effectively exploits systematic disruptions in compressed byte patterns without ever recovering the original text.
Extensive evaluations confirm its superiority: it achieves a state-of-the-art average F1-score of 0.9909 across five benchmarks, outperforming the best baseline by 2.72 percentage points while operating exclusively in the compressed domain.

\begin{acks}
This work is supported by the ***.
\end{acks}


\bibliographystyle{ACM-Reference-Format}
\bibliography{compression}

@String{Computing = "Computing" }

@String{Computer = "{IEEE} Computer" }

@String{Springer = "Springer-Verlag" }

@String{VanNostrand = "Van Nostrand" }

@inproceedings{wang2024muslope,
  title={$\mu$Slope: High Compression and Fast Search on Semi-Structured Logs},
  author={Wang, Rui and Gibson, Devin and Rodrigues, Kirk and Luo, Yu and Zhang, Yun and Wang, Kaibo and Fu, Yupeng and Chen, Ting and Yuan, Ding},
  booktitle={18th USENIX Symposium on Operating Systems Design and Implementation (OSDI 24)},
  pages={529--544},
  year={2024}
}

@inproceedings{yu2023logreducer,
  title={Logreducer: Identify and reduce log hotspots in kernel on the fly},
  author={Yu, Guangba and Chen, Pengfei and Li, Pairui and Weng, Tianjun and Zheng, Haibing and Deng, Yuetang and Zheng, Zibin},
  booktitle={2023 IEEE/ACM 45th International Conference on Software Engineering (ICSE)},
  pages={1763--1775},
  year={2023},
  organization={IEEE}
}

@inproceedings{liu2019logzip,
  title={Logzip: Extracting hidden structures via iterative clustering for log compression},
  author={Liu, Jinyang and Zhu, Jieming and He, Shilin and He, Pinjia and Zheng, Zibin and Lyu, Michael R},
  booktitle={2019 34th IEEE/ACM International Conference on Automated Software Engineering (ASE)},
  pages={863--873},
  year={2019},
  organization={IEEE}
}

@inproceedings{rodrigues2021clp,
  title={CLP: Efficient and scalable search on compressed text logs},
  author={Rodrigues, Kirk and Luo, Yu and Yuan, Ding},
  booktitle={15th USENIX Symposium on Operating Systems Design and Implementation (OSDI 21)},
  pages={183--198},
  year={2021}
}

@inproceedings{wei2023loggrep,
  title={Loggrep: Fast and cheap cloud log storage by exploiting both static and runtime patterns},
  author={Wei, Junyu and Zhang, Guangyan and Chen, Junchao and Wang, Yang and Zheng, Weimin and Sun, Tingtao and Wu, Jiesheng and Jiang, Jiangwei},
  booktitle={Proceedings of the Eighteenth European Conference on Computer Systems},
  pages={452--468},
  year={2023}
}

@inproceedings{wei2021logreducer,
  title={On the feasibility of parser-based log compression in Large-Scale cloud systems},
  author={Wei, Junyu and Zhang, Guangyan and Wang, Yang and Liu, Zhiwei and Zhu, Zhanyang and Chen, Junchao and Sun, Tingtao and Zhou, Qi},
  booktitle={19th USENIX Conference on File and Storage Technologies (FAST 21)},
  pages={249--262},
  year={2021}
}

@article{tang2025loglite,
  title={LogLite: Lightweight Plug-and-Play Streaming Log Compression},
  author={Tang, Benzhao and Yang, Shiyu and Shen, Zhitao and Zhang, Wenjie and Lin, Xuemin and Tian, Zhihong},
  journal={Proceedings of the VLDB Endowment},
  volume={18},
  number={11},
  pages={3757--3770},
  year={2025},
  publisher={VLDB Endowment}
}

@inproceedings{du2017deeplog,
  title={Deeplog: Anomaly detection and diagnosis from system logs through deep learning},
  author={Du, Min and Li, Feifei and Zheng, Guineng and Srikumar, Vivek},
  booktitle={Proceedings of the 2017 ACM SIGSAC conference on computer and communications security},
  pages={1285--1298},
  year={2017}
}

@inproceedings{zhang2019robust,
  title={Robust log-based anomaly detection on unstable log data},
  author={Zhang, Xu and Xu, Yong and Lin, Qingwei and Qiao, Bo and Zhang, Hongyu and Dang, Yingnong and Xie, Chunyu and Yang, Xinsheng and Cheng, Qian and Li, Ze and others},
  booktitle={Proceedings of the 2019 27th ACM joint meeting on European software engineering conference and symposium on the foundations of software engineering},
  pages={807--817},
  year={2019}
}

@inproceedings{xie2022loggd,
  title={Loggd: Detecting anomalies from system logs with graph neural networks},
  author={Xie, Yongzheng and Zhang, Hongyu and Babar, Muhammad Ali},
  booktitle={2022 IEEE 22nd International conference on software quality, reliability and security (QRS)},
  pages={299--310},
  year={2022},
  organization={IEEE}
}

@inproceedings{le2021log,
  title={Log-based anomaly detection without log parsing},
  author={Le, Van-Hoang and Zhang, Hongyu},
  booktitle={2021 36th IEEE/ACM International Conference on Automated Software Engineering (ASE)},
  pages={492--504},
  year={2021},
  organization={IEEE}
}

@inproceedings{guo2021logbert,
  title={Logbert: Log anomaly detection via bert},
  author={Guo, Haixuan and Yuan, Shuhan and Wu, Xintao},
  booktitle={2021 international joint conference on neural networks (IJCNN)},
  pages={1--8},
  year={2021},
  organization={IEEE}
}

@inproceedings{zhou2024f,
  title={F-tadoc: Fpga-based text analytics directly on compression with hls},
  author={Zhou, Yanliang and Zhang, Feng and Lin, Tuo and Huang, Yuanjie and Long, Saiqin and Zhai, Jidong and Du, Xiaoyong},
  booktitle={2024 IEEE 40th International Conference on Data Engineering (ICDE)},
  pages={3739--3752},
  year={2024},
  organization={IEEE}
}

@inproceedings{zhang2022compressdb,
  title={CompressDB: Enabling efficient compressed data direct processing for various databases},
  author={Zhang, Feng and Wan, Weitao and Zhang, Chenyang and Zhai, Jidong and Chai, Yunpeng and Li, Haixiang and Du, Xiaoyong},
  booktitle={Proceedings of the 2022 International Conference on Management of Data},
  pages={1655--1669},
  year={2022}
}

@article{guan2023homomorphic,
  title={Homomorphic compression: Making text processing on compression unlimited},
  author={Guan, Jiawei and Zhang, Feng and Ma, Siqi and Chen, Kuangyu and Hu, Yihua and Chen, Yuxing and Pan, Anqun and Du, Xiaoyong},
  journal={Proceedings of the ACM on Management of Data},
  volume={1},
  number={4},
  pages={1--28},
  year={2023},
  publisher={ACM New York, NY, USA}
}

@inproceedings{lu2018detecting,
  title={Detecting anomaly in big data system logs using convolutional neural network},
  author={Lu, Siyang and Wei, Xiang and Li, Yandong and Wang, Liqiang},
  booktitle={2018 IEEE 16th Intl Conf on Dependable, Autonomic and Secure Computing, 16th Intl Conf on Pervasive Intelligence and Computing, 4th Intl Conf on Big Data Intelligence and Computing and Cyber Science and Technology Congress (DASC/PiCom/DataCom/CyberSciTech)},
  pages={151--158},
  year={2018},
  organization={IEEE}
}

@inproceedings{Loghub2023,
  author       = {Jieming Zhu and
                  Shilin He and
                  Pinjia He and
                  Jinyang Liu and
                  Michael R. Lyu},
  title        = {Loghub: {A} Large Collection of System Log Datasets for AI-driven 
                  Log Analytics},
  booktitle    = {IEEE International Symposium on Software Reliability Engineering (ISSRE)},
  year         = {2023}
}

@inproceedings{oliner2007supercomputers,
  title={What supercomputers say: A study of five system logs},
  author={Oliner, Adam and Stearley, Jon},
  booktitle={37th annual IEEE/IFIP international conference on dependable systems and networks (DSN'07)},
  pages={575--584},
  year={2007},
  organization={IEEE}
}

@inproceedings{xu2009online,
  title={Online system problem detection by mining patterns of console logs},
  author={Xu, Wei and Huang, Ling and Fox, Armando and Patterson, David and Jordan, Michael},
  booktitle={2009 ninth IEEE international conference on data mining},
  pages={588--597},
  year={2009},
  organization={IEEE}
}

@inproceedings{meng2019loganomaly,
  title={Loganomaly: Unsupervised detection of sequential and quantitative anomalies in unstructured logs.},
  author={Meng, Weibin and Liu, Ying and Zhu, Yichen and Zhang, Shenglin and Pei, Dan and Liu, Yuqing and Chen, Yihao and Zhang, Ruizhi and Tao, Shimin and Sun, Pei and others},
  booktitle={Ijcai},
  volume={19},
  number={7},
  pages={4739--4745},
  year={2019}
}

@inproceedings{yang2021semi,
  title={Semi-supervised log-based anomaly detection via probabilistic label estimation},
  author={Yang, Lin and Chen, Junjie and Wang, Zan and Wang, Weijing and Jiang, Jiajun and Dong, Xuyuan and Zhang, Wenbin},
  booktitle={2021 IEEE/ACM 43rd International Conference on Software Engineering (ICSE)},
  pages={1448--1460},
  year={2021},
  organization={IEEE}
}

@inproceedings{wan2021glad,
  title={Glad-paw: Graph-based log anomaly detection by position aware weighted graph attention network},
  author={Wan, Yi and Liu, Yilin and Wang, Dong and Wen, Yujin},
  booktitle={Pacific-asia conference on knowledge discovery and data mining},
  pages={66--77},
  year={2021},
  organization={Springer}
}

@inproceedings{qi2023loggpt,
  title={Loggpt: Exploring chatgpt for log-based anomaly detection},
  author={Qi, Jiaxing and Huang, Shaohan and Luan, Zhongzhi and Yang, Shu and Fung, Carol and Yang, Hailong and Qian, Depei and Shang, Jing and Xiao, Zhiwen and Wu, Zhihui},
  booktitle={2023 IEEE international conference on high performance computing \& communications, data science \& systems, smart city \& dependability in sensor, cloud \& big data systems \& application (HPCC/DSS/SmartCity/DependSys)},
  pages={273--280},
  year={2023},
  organization={IEEE}
}

@inproceedings{sui2025bridging,
  title={Bridging the gap: Llm-powered transfer learning for log anomaly detection in new software systems},
  author={Sui, Yicheng and Wang, Xiaotian and Cui, Tianyu and Xiao, Tong and He, Chenghao and Zhang, Shenglin and Zhang, Yuzhi and Yang, Xiao and Sun, Yongqian and Pei, Dan},
  booktitle={2025 IEEE 41st International Conference on Data Engineering (ICDE)},
  pages={4414--4427},
  year={2025},
  organization={IEEE}
}

@article{zhu2025cola,
  title={CoLA: Model Collaboration for Log-based Anomaly Detection},
  author={Zhu, Xuhang and Tang, Xiu and Wu, Sai and Li, Jichen and Wang, Haobo and Yao, Chang and Xu, Quanqing and Chen, Gang},
  journal={Proceedings of the VLDB Endowment},
  volume={18},
  number={11},
  pages={3979--3987},
  year={2025},
  publisher={VLDB Endowment}
}

@article{wang2025logcioud,
  title={LogCIoud: Fast Search of Compressed Logs on Object Storage},
  author={Wang, Ziheng and Wei, Junyu and Aiken, Alex and Zhang, Guangyan and T{\o}rring, Jacob O and Jiang, Rain and Jiang, Chenyu and Xu, Wei},
  journal={Proceedings of the VLDB Endowment},
  volume={18},
  number={8},
  pages={2362--2370},
  year={2025},
  publisher={VLDB Endowment}
}

@article{fan2025inference,
  title={Inference-Friendly Graph Compression for Graph Neural Networks},
  author={Fan, Yangxin and Che, Haolai and Wu, Yinghui},
  journal={Proceedings of the VLDB Endowment},
  volume={18},
  number={9},
  pages={3203--3215},
  year={2025},
  publisher={VLDB Endowment}
}

@inproceedings{wu2018group,
  title={Group normalization},
  author={Wu, Yuxin and He, Kaiming},
  booktitle={Proceedings of the European conference on computer vision (ECCV)},
  pages={3--19},
  year={2018}
}

@article{vaswani2017attention,
  title={Attention is all you need},
  author={Vaswani, Ashish and Shazeer, Noam and Parmar, Niki and Uszkoreit, Jakob and Jones, Llion and Gomez, Aidan N and Kaiser, {\L}ukasz and Polosukhin, Illia},
  journal={Advances in neural information processing systems},
  volume={30},
  year={2017}
}

@inproceedings{xiong2020layer,
  title={On layer normalization in the transformer architecture},
  author={Xiong, Ruibin and Yang, Yunchang and He, Di and Zheng, Kai and Zheng, Shuxin and Xing, Chen and Zhang, Huishuai and Lan, Yanyan and Wang, Liwei and Liu, Tieyan},
  booktitle={International conference on machine learning},
  pages={10524--10533},
  year={2020},
  organization={PMLR}
}

@article{zhang2019root,
  title={Root mean square layer normalization},
  author={Zhang, Biao and Sennrich, Rico},
  journal={Advances in neural information processing systems},
  volume={32},
  year={2019}
}

@article{shazeer2020glu,
  title={Glu variants improve transformer},
  author={Shazeer, Noam},
  journal={arXiv preprint arXiv:2002.05202},
  year={2020}
}

@article{beck2024xlstm,
  title={xlstm: Extended long short-term memory},
  author={Beck, Maximilian and P{\"o}ppel, Korbinian and Spanring, Markus and Auer, Andreas and Prudnikova, Oleksandra and Kopp, Michael and Klambauer, G{\"u}nter and Brandstetter, Johannes and Hochreiter, Sepp},
  journal={Advances in Neural Information Processing Systems},
  volume={37},
  pages={107547--107603},
  year={2024}
}

@article{oord2018representation,
  title={Representation learning with contrastive predictive coding},
  author={Oord, Aaron van den and Li, Yazhe and Vinyals, Oriol},
  journal={arXiv preprint arXiv:1807.03748},
  year={2018}
}

@inproceedings{lin2017focal,
  title={Focal loss for dense object detection},
  author={Lin, Tsung-Yi and Goyal, Priya and Girshick, Ross and He, Kaiming and Doll{\'a}r, Piotr},
  booktitle={Proceedings of the IEEE international conference on computer vision},
  pages={2980--2988},
  year={2017}
}

@article{khosla2020supervised,
  title={Supervised contrastive learning},
  author={Khosla, Prannay and Teterwak, Piotr and Wang, Chen and Sarna, Aaron and Tian, Yonglong and Isola, Phillip and Maschinot, Aaron and Liu, Ce and Krishnan, Dilip},
  journal={Advances in neural information processing systems},
  volume={33},
  pages={18661--18673},
  year={2020}
}

@article{loshchilov2017decoupled,
  title={Decoupled weight decay regularization},
  author={Loshchilov, Ilya and Hutter, Frank},
  journal={arXiv preprint arXiv:1711.05101},
  year={2017}
}

@article{ma2024pluto,
  title={Pluto: Sample selection for robust anomaly detection on polluted log data},
  author={Ma, Lei and Cao, Lei and VanNostrand, Peter M and Hofmann, Dennis M and Su, Yao and Rundensteiner, Elke A},
  journal={Proceedings of the ACM on Management of Data},
  volume={2},
  number={4},
  pages={1--25},
  year={2024},
  publisher={ACM New York, NY, USA}
}

@article{jia2023robust,
  title={Robust and transferable log-based anomaly detection},
  author={Jia, Peng and Cai, Shaofeng and Ooi, Beng Chin and Wang, Pinghui and Xiong, Yiyuan},
  journal={Proceedings of the ACM on Management of Data},
  volume={1},
  number={1},
  pages={1--26},
  year={2023},
  publisher={ACM New York, NY, USA}
}

@inproceedings{chen2020logtransfer,
  title={Logtransfer: Cross-system log anomaly detection for software systems with transfer learning},
  author={Chen, Rui and Zhang, Shenglin and Li, Dongwen and Zhang, Yuzhe and Guo, Fangrui and Meng, Weibin and Pei, Dan and Zhang, Yuzhi and Chen, Xu and Liu, Yuqing},
  booktitle={2020 IEEE 31st International Symposium on Software Reliability Engineering (ISSRE)},
  pages={37--47},
  year={2020},
  organization={IEEE}
}

@inproceedings{zhang2024multivariate,
  title={Multivariate log-based anomaly detection for distributed database},
  author={Zhang, Lingzhe and Jia, Tong and Jia, Mengxi and Li, Ying and Yang, Yong and Wu, Zhonghai},
  booktitle={Proceedings of the 30th ACM SIGKDD Conference on Knowledge Discovery and Data Mining},
  pages={4256--4267},
  year={2024}
}

@inproceedings{liu2024tp,
  title={Tp-gnn: Continuous dynamic graph neural network for graph classification},
  author={Liu, Jie and Liu, Jiamou and Zhao, Kaiqi and Tang, Yanni and Chen, Wu},
  booktitle={2024 IEEE 40th International Conference on Data Engineering (ICDE)},
  pages={2848--2861},
  year={2024},
  organization={IEEE}
}

@article{tang2024substructure,
  title={Substructure-Aware Log Anomaly Detection},
  author={Tang, Yanni and Zhang, Zhuoxing and Zhao, Kaiqi and Fang, Lanting and Li, Zhenhua and Chen, Wu},
  journal={Proceedings of the VLDB Endowment},
  volume={18},
  number={2},
  pages={213--225},
  year={2024},
  publisher={VLDB Endowment}
}

@inproceedings{hu2025cost,
  title={A cost-effective and decompression-transparent compressor for OLTP-oriented databases},
  author={Hu, Hao and Zheng, Qiyang and Zou, Xiangyu and Qin, Lisha and Zhang, Chengwei and Zhang, Wanchuan and Jiang, Zhaoheng and Tao, Dingwen and Wang, Hongpeng and Xia, Wen},
  booktitle={2025 IEEE 41st International Conference on Data Engineering (ICDE)},
  pages={405--418},
  year={2025},
  organization={IEEE}
}

\end{document}